\pdfoutput=1

\documentclass[11pt]{article}

\usepackage{acl}

\usepackage{times}
\usepackage{latexsym}

\usepackage[T1]{fontenc}

\usepackage[utf8]{inputenc}

\usepackage{microtype}

\usepackage{inconsolata}

\usepackage{graphicx}

\usepackage{hyperref}

\usepackage{amsmath}
\usepackage{amssymb}
\usepackage{mathtools}
\usepackage{amsthm}

\usepackage{kotex}
\usepackage{multirow}
\usepackage{longtable}
\usepackage[normalem]{ulem}
\useunder{\uline}{\ul}{}
\usepackage{array}
\usepackage{caption}

\title{ESG-Kor: A Korean Dataset for ESG-related Information Extraction and Practical Use Cases}

\author{
  \textbf{Jaeyoung Lee}\quad
  \textbf{Geonyeong Son}\quad
  \textbf{Misuk Kim\thanks{Corresponding Author}}
\\
  Department of Data Science, Hanyang University, South Korea
\\
  $\left\{\texttt{jaeylee, geonyeongson, misukkim}\right\}$\texttt{@hanyang.ac.kr}
  }

\begin{document}
\maketitle
\begin{abstract}
With the expansion of pre-trained language model usage in recent years, the importance of datasets for performing tasks in specialized domains has significantly increased. Therefore, we have built a Korean dataset called \texttt{ESG-Kor} to automatically extract Environmental, Social, and Governance (ESG) information, which has recently gained importance. \texttt{ESG-Kor} is a dataset consisting of a total of 118,946 sentences that extracted information on each ESG component from Korean companies' sustainability reports and manually labeled it according to objective rules provided by ESG evaluation agencies. To verify the effectiveness and applicability of the \texttt{ESG-Kor} dataset, classification performance was confirmed using several Korean pre-trained language models, and significant performance was obtained. Additionally, by extending the ESG classification model to documents of small and medium enterprises and extracting information based on ESG key issues and in-depth analysis, we demonstrated potential and practical use cases in the ESG field.
\end{abstract}

\section{Introduction}
Pre-trained language models are exhibiting astonishing performances in various natural language processing (NLP) tasks, including classification, question answering (QA), machine translation, summarization, and conversation generation, and are currently being utilized in numerous fields. Moreover, pre-trained language models specialized domains, such as FinBERT \cite{yang2020finbert}, Med-BERT \cite{rasmy2021med}, and LEGAL-BERT \cite{chalkidis2020legal}, which train the text of the respective domains, have emerged and shown superior performance in solving problems of each domain \cite{gururanga2020dont}. However, these pre-trained language models show their limits when applied to more detailed subfields, leading to an increased significance of datasets from such detailed subfields \cite{hendrycks2021cuad}. Despite this, the construction of datasets from these detailed subfields necessitates the involvement of relevant experts, thereby requiring more time and costs and resulting in limited construction of related datasets.
Moreover, most publicly available datasets and pre-trained language models are built based on English, limiting their applicability for performing specific tasks in non-English languages. While there have been pre-trained language models released for supporting multiple languages \cite{pires2019multilingual}, they do not perform as well as the models based on a single language. Consequently, studies are being conducted to build datasets for specific language-based tasks by translating pre-constructed English data into an other language \cite{park2021klue}. Nevertheless, these exhibit limited performance compared to English-based pre-trained language models and are especially difficult to use in highly specialized domains, prompting ongoing research into constructing datasets for specialized domains in diverse languages \cite{zadeh2020cmu}.

In this paper, we introduce \texttt{ESG-Kor}, a large-scale Korean dataset that encompasses Environmental, Social, and Governance (ESG) information. Considering the importance of ethical management, social issues such as racial and gender discrimination, and environmental problems like climate change, ESG is essential for a sustainable society. The importance of ESG is increasing globally as a measure of non-financial corporate activities. ESG information is mainly provided as text without a standardized format, making it very challenging to obtain only the necessary ESG information from a vast amount of data \cite{lee2023esg}, and small-scale companies do not provide reports for the disclosure of ESG information. Therefore, the potential benefits are significant if ESG information can be extracted from various text data through an ESG-based language model. For this, we used sustainability reports related to ESG in this study to build a Korean ESG dataset. We constructed a dataset of 118,946 sentences, manually labeled by three annotators possessing relevant knowledge. The annotators were conducted based on objective rules provided by ESG rating providers. We verified the performance by fine-tuning the existing pre-trained language models with the constructed \texttt{ESG-Kor} dataset and validated the effectiveness and applicability of \texttt{ESG-Kor}.

Additionally, by extending our methodology and utilizing additional techniques, we demonstrate potential and practical use cases, such as the expansion of ESG into the small and medium enterprises (SMEs) environment and in-depth analysis of corporate ESG based on ESG key issues. Our contributions are as follows:
\begin{itemize}
    \item We constructed \texttt{ESG-Kor}, a large-scale dataset for ESG sentence extraction and classification, comprised of 118,946 sentences by manually labeling sentences from sustainability reports, which are documents related to ESG. \texttt{ESG-Kor} is the first constructed large-scale dataset associated with ESG.
    \item We fine-tuned pre-trained language models using the \texttt{ESG-Kor}. Through experiments, we confirmed that the model appropriately performs the tasks of ESG sentence extraction and classification, verifying the dataset's usefulness.
    \item By extending our approach, we presented two use cases: the expansion of ESG into the SME environment and the extraction of ESG key issues. Through these use cases, we demonstrated the potential and usefulness of our dataset and model in the ESG field.
\end{itemize}

\begin{figure*}[t!]
\vskip 0.1in
\centering
\includegraphics[width=0.9\linewidth]{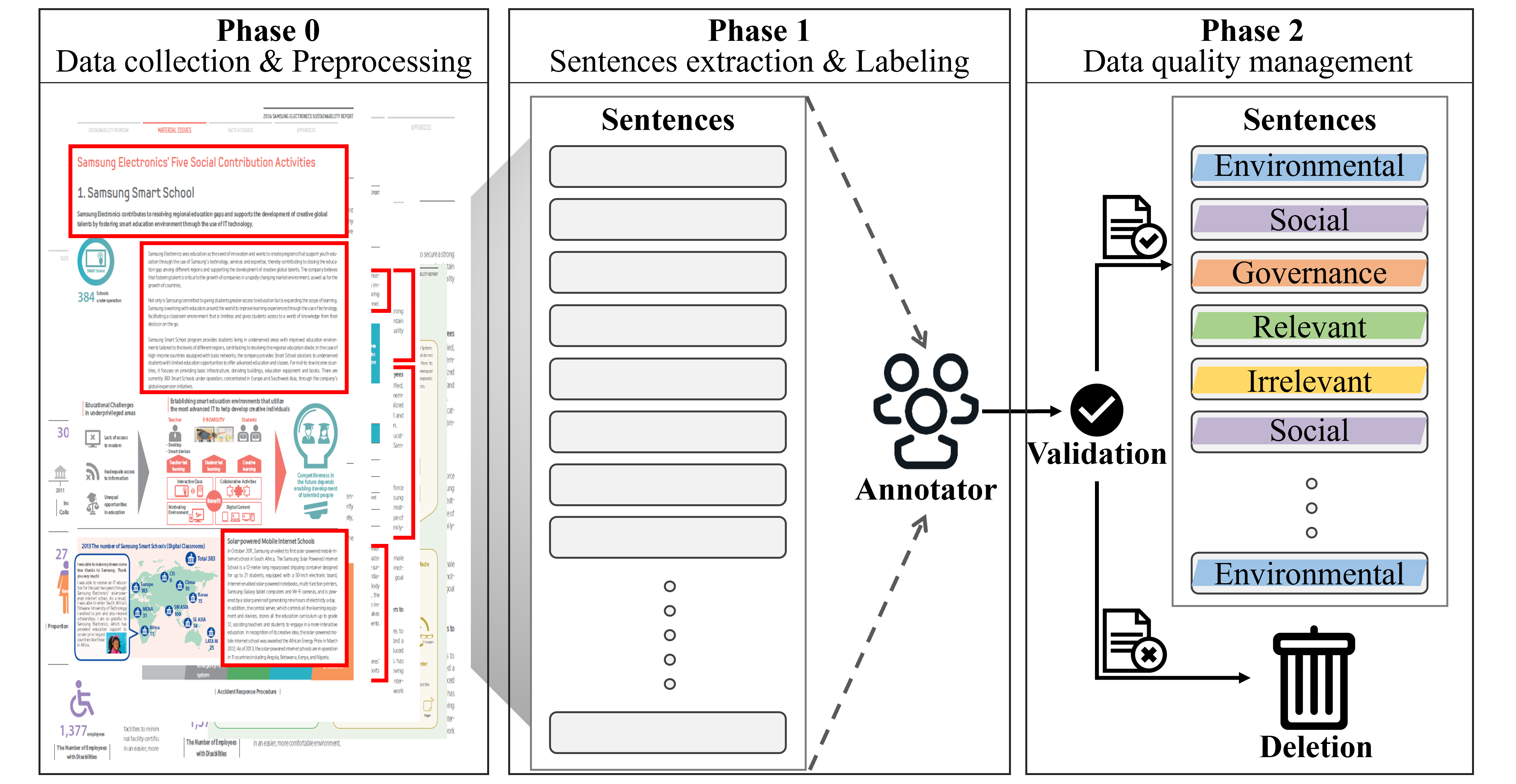}
\caption{\texttt{ESG-Kor} construction process. Phase 0: Collect sustainability reports data and undergo preprocessing. Phase 1: Extract only the text from the preprocessed data, and proceed with labeling through annotators equipped with ESG-relevant knowledge. Phase 2: For the data quality, go through a verification process, and sentences of low quality are deleted. The example of the sustainability report on the left side of the figure is from the 2014 sustainability report published by Samsung Electronics.}
\label{ESG_kor_ex}
\vskip -0.1in
\end{figure*}
\texttt{ESG-Kor} is the only Korean-based professional NLP benchmark dataset manually labeled in the field of ESG. The entire construction process of \texttt{ESG-Kor} is illustrated in Figure \ref{ESG_kor_ex}. This dataset can be leveraged by diverse stakeholders, and the data construction process can be used as a benchmark for constructing ESG datasets in various languages. We release the \texttt{ESG-Kor} dataset using a data repository to advance ESG research. We have created temporary accounts with guaranteed anonymity to provide some dataset samples. The dataset can be accessed at \url{https://github.com/nowzer0/ESG-Kor}.

\section{Related Work}
For pre-trained language models to be effectively utilized in diverse nations and highly specialized domains, datasets and related research tailored to different languages and specialized domains are essential. Moreover, despite the increasing importance of ESG, research leverages machine learning techniques in the field of ESG is limited \cite{macpherson2021implications}. We present research focused on specialized domains and minority languages, as well as research related to ESG.

\subsection{Specialized Domain \&\ Minority Language}
Datasets tailored to specialized domains are being built. First, in the biomedical domain, a Chinese-based biomedical language understanding benchmark \cite{zhang2022cblue}, the PubMedQA \cite{jin2019pubmedqa}, which is a biomedical QA set processed from data collected from the medical information search engine PubMed, the comprehensive benchmark dataset for PubMed-based Biomedical NLP known as BLURB \cite{gu2021domain}, and the CORD-19 public research dataset provided for rapid access to COVID-19 researchers \cite{wang2020cord} have been constructed. In the legal domain, LexGLUE \cite{chalkidis2022lexglue}, which standardizes various tasks for evaluation, and CUAD \cite{hendrycks2021cuad}, a dataset with annotations added by legal experts over a long period for reviewing legal contracts, have been released. In addition, \citet{li2023geoglue} introduced the GeoGLUE dataset specialized for geographic understanding by collecting data from publicly available geographic resources for six natural language understanding tasks. The MultiHIERTT dataset \cite{zhao2022multihiertt} has also been suggested for inferring answers to finance-related questions.

Furthermore, datasets based on diverse languages are being constructed. \citet{yum2021word} have developed a word embedding dataset that captures semantic similarity and relatedness specifically for Korean medical terms. For Japanese-based datasets, NWJC2Vec \cite{asahara2018nwjc2vec}, a Japanese word embedding dataset, JaQuAD \cite{so2022jaquad}, a QA dataset using Japanese Wikipedia data, and Japanese clinical domain semantic textual similarity \cite{mutinda2021semantic} have been created. Datasets constructed in Russian include RuBQ \cite{korablinov2020rubq}, a knowledge-based QA dataset through Wikidata, SberQuAD \cite{efimov2020sberquad}, a large-scale Reading Comprehension (RC) dataset created similarly to English-based SQuAD \cite{rajpurka2016squad}, and NEREL-BIO \cite{loukachevitch2023nerel}, a PubMed abstract dataset. For Spanish datasets, there is the Biomedical dataset CoWeSe \cite{carrino2021spanish} and a word embedding dataset in the Medical domain \cite{soares2019medical}. In the realm of Arabic datasets, notable examples include the dialect dataset introduced by \citet{abdul2018you} and the aspect-based sentiment analysis dataset presented by \citet{obiedat2021arabic}. In addition, research efforts focus on creating publicly available resources in various areas. For instance, CMU-MOSEAS \cite{zadeh2020cmu} dataset, a multi-modal sentiment dataset, has been made accessible, which encompasses not only a single language but also provides broader multilingual coverage, including languages such as Spanish, Portuguese, German, and French. Hence, constructing datasets not just for specific domains but for diverse languages is a crucial task.

\subsection{ESG Research using Machine Learning}
Unlike the traditional approach of evaluating companies based on their financial performance, the focus on non-financial performance based on sustainability has increased, leading to increasing interest in ESG. Accordingly, machine learning research is being conducted in the field of ESG to solve various problems, but it is limited due to the lack of expertise or relevant ESG datasets. \citet{de2020does} used ESG indicators and economic indicators to predict the correlation between the two indicators and key financial indicators through various machine learning algorithms. \citet{lanza2020mind} identified ESG indicators effective for building an investment portfolio through tree algorithms.  Recently, research using ESG ratings provided by ESG evaluation companies is also being conducted \cite{de2020esg, serafeim2022corporate}. With the emergence of high-performance pre-trained language models, research has also been conducted to classify sentences as sustainable or unsustainable using pre-trained language models \cite{pontes2022using}, build an ESG index through media data \cite{sokolov2020building}, profile ESG-related language in discourse, and construct an ESG relevance dataset through a similarity-based unsupervised method \cite{raman2020mapping}. Machine learning in the ESG field can solve various ESG problems and is expected to reduce considerable environmental and social costs, so constructing ESG-related datasets is necessary. In the environmental aspect of ESG, recent research has been published on constructing datasets for detecting the presence of Net zero and reduction targets in texts provided by various organizations \cite{schimanski2023climatebert}. In the ESG field, machine learning is needed to address multiple ESG issues. It can be expected to contribute to solving sustainable society challenges, offering significant environmental and social cost reduction benefits. Therefore, the construction of ESG-related datasets is essential.

\section{\texttt{ESG-K}\texttt{\lowercase{or}} dataset}
\subsection{Sustainability Reports}
We used Korean companies' sustainability reports as raw data for constructing \texttt{ESG-Kor} dataset. A sustainability report is a report that discloses a company's significant sustainability impacts on stakeholders based on economic, social, and environmental activities, as well as the company's performance over the past year. It reports various non-financial performances and includes various ESG-related information. Many diverse companies are publishing sustainability reports due to the increased demand for disclosure of ESG-related information from various company stakeholders and the institutionalization of the sustainability report disclosure system \cite{siew2015review}.

We have collected sustainability reports from 53 sectors of Korean companies published from 2002 to 2021. Since most sectors have fewer than 20 reports or have a small amount of data in the reports, we initially selected the automobile, aviation, electronics, finance, and heavy industry sectors, which had relatively more data among the sectors with more than 20 published reports, to use as raw data. For the purpose of evaluating data utility, we also built a dataset from the sustainability reports of \textsf{POSCO} in the steel sector and \textsf{NAVER} in the IT sector, which are significant companies in Korea.

\begin{table}[h!]
\centering
\vskip 0.1in
\resizebox{\columnwidth}{!}{
\begin{tabular}{llcc}
\hline
\multicolumn{1}{c}{\multirow{2}{*}{Sector}}         & \multicolumn{1}{c}{\multirow{2}{*}{Company}} & Number of & Number of \\
                                &                          & reports   & sentences \\ \hline
\multirow{2}{*}{Automobile}     & Hyundai motor company    & 17        & 15,154     \\
                                & KIA corporation          & 14        & 9,921      \\ \hline
\multirow{2}{*}{Aviation}       & Korean airlines          & 17        & 7,022      \\
                                & Asiana airlines          & 8         & 2,302      \\ \hline
\multirow{2}{*}{Electronics}    & Samsung electronics      & 14        & 12,491     \\
                                & LG electronics           & 14        & 11,107     \\ \hline
\multirow{6}{*}{Finance}        & BNK financial group      & 7         & 4,591      \\
                                & Hana financial group     & 10        & 5,080      \\
                                & Industrial bank of Korea & 6         & 3,435      \\
                                & KB financial group       & 10        & 5,837      \\
                                & Shinhan financial group  & 14        & 8,091      \\
                                & Woori financial group    & 1         & 707       \\ \hline
\multirow{3}{*}{Heavy industry} & Hyundai heavy industries & 6         & 3,420      \\
                                & Samsung heavy industries & 10        & 5,219      \\
                                & Doosan heavy industries  & 8         & 4,970      \\ \hline
Steel industry                  & POSCO                    & 16        & 18,886     \\ \hline
IT                              & NAVER                    & 1         & 713       \\ \hline
\multicolumn{2}{c}{Total}                                  & 173       & 118,946    \\ \hline
\end{tabular}%
}
\caption{Sustainability reports data}
\label{number of reports}
\vskip -0.1in
\end{table}
The sustainability reports used to build the dataset consist of 173 volumes, and the number of reports per sector is presented in Table \ref{number of reports}. For \textsf{NAVER}, only one report had been published as of 2021, and only this report was used.

\subsection{Preprocessing}
The collected 173 volumes of sustainability reports are provided in portable document format (PDF) and contain various forms of data, including text, tables, and images. We extracted only the text from the PDF reports for dataset construction. To label each sentence, we used the Korean sentence splitter (KSS) \cite{ko2021kss}, a Korean sentence separation library, to distinguish the extracted text data at the sentence level. For ease of analysis, we performed preprocessing tasks such as unifying the numerical data format, deleting special characters, and removing duplicate sentences, and used the Korean spell-check API, py-hanspell \cite{pyhanspell}, to perform spell checks. Finally, we manually check the noise data generated by automatically not properly divided sentences, or too long or too short sentences, and perform segmentation or deletion.
 
\subsection{Labeling}
We manually labeled the preprocessed sustainability reports to construct \texttt{ESG-Kor} for extracting and classifying ESG information. The labeling was referenced from the criteria of leading global ESG rating providers and Korean ESG evaluation agencies such as Morgan Stanley Capital International (MSCI), Refinitiv, and the Korean Institute of Corporate Governance and Sustainability (KCGS). Three annotators, equipped with relevant knowledge based on the standards of each institution, familiarized and organized the labeling rules. The labeling rules are as follows:
\begin{itemize}
\item {\verb|E(Environmental)|} Environmental factors such as green management, climate change, carbon emissions, natural resources, waste, resource use, energy conservation, etc.
\item {\verb|S(Social)|} Social factors such as human capital, community service, workers, working environment, suppliers, community contribution, human rights, gender and diversity, customer satisfaction, information protection, etc.
\item {\verb|G(Governance)|} Governance factors such as ethical management, shareholder rights, board of directors, audit organization, stakeholders, dividends, regulatory compliance, taxation, etc.
\item {\verb|R(Relevant)|} Containing ESG-related information but not clearly categorized into individual ESG elements or including contents of both ESG elements simultaneously,
\item {\verb|I(Irrelevant)|}: Not including ESG information.
\end{itemize}
The labeling was divided into two stages to ensure the dataset's quality. First, three annotators well-acquainted with the labeling rules annotated sentences clearly classified into each class. Subsequently, hard sentences to distinguish were separately discussed among the three annotators. They annotated if all agreed on labeling it to a specific class. However, if even one annotator had a different opinion, the sentence was deleted to prevent potential degradation of dataset quality. Through discussions among annotators, the majority of sentences were annotated into respective classes, and sentences deleted due to differing opinions among annotators, comprising less than 0.1\% of the total dataset. The dataset construction involved labeling for 2 hours a day, five days a week, with an average of 500-600 sentences labeled per 2 hours. Therefore, each annotator labeled approximately 2,500 sentences per week, and the process first focused on labeling clear sentences over about four months. In the following two months, sentences that were difficult to categorize were discussed for labeling or deletion, and a period of cross-checking was conducted over another two months, totaling approximately 8-9 months of labeling time. Examples of sentences for each class in the \texttt{ESG-Kor} dataset are shown in Table \ref{ex_sen}.

\begin{table}[h!]
\centering
\vskip 0.1in
\resizebox{\columnwidth}{!}{%
\begin{tabular}{cl}
\hline
Class &
  \multicolumn{1}{c}{Sentences} \\ \hline
E &
  \begin{tabular}[c]{@{}l@{}}We consider the environmental impact of new products \\ during their whole development cycle.\end{tabular} \\ \hline
S &
  \begin{tabular}[c]{@{}l@{}}The company also has created open communication \\ channels to maintain strong relationships with suppliers.\end{tabular} \\ \hline
G &
  \begin{tabular}[c]{@{}l@{}}Manages a website that offers information on ethical \\business management and provides an external reporting \\mechanism since 2002.\end{tabular} \\ \hline
R &
  \begin{tabular}[c]{@{}l@{}}Analyzing the social and environmental impact of their\\ manufacturing process and communicate the company’s\\ efforts with major suppliers in a transparent manner - \\something the company hopes to accomplish through\\ sustainability reporting.\end{tabular} \\ \hline
I &
  \begin{tabular}[c]{@{}l@{}}The company has also introduced a new focus on the \\business to business (B2B) market, going beyond the \\business-to-consumer (B2C) market\end{tabular} \\ \hline
\end{tabular}%
}
\caption{Examples of labeled sentences}
\label{ex_sen}
\vskip -0.1in
\end{table}
Also, the number of sentences by sector for each ESG element in the dataset is shown in Table \ref{datasets_statics}.

\begin{table*}[h!]
\centering
\vskip 0.1in
\resizebox{0.8\textwidth}{!}{%
\begin{tabular}{ccccccccc}
\hline
Class &
  \begin{tabular}[c]{@{}c@{}}Automobile\end{tabular} &
  \begin{tabular}[c]{@{}c@{}}Aviation\end{tabular} &
  \begin{tabular}[c]{@{}c@{}}Electronics\end{tabular} &
  \begin{tabular}[c]{@{}c@{}}Finance\end{tabular} &
  \begin{tabular}[c]{@{}c@{}}Heavy industry\end{tabular} &
  \begin{tabular}[c]{@{}c@{}}Steel industy\end{tabular} &
  IT &
  Total \\ \hline
E     & 5,783  & 1,727 & 4,511  & 2,228  & 1,779  & 3,701  & 60  & 19,789  \\
S     & 8,466  & 3,946 & 9,033  & 12,578 & 5,693  & 5,802  & 264 & 45,782  \\
G     & 896    & 463   & 1,928  & 3,014  & 751    & 1,690  & 77  & 8,819   \\
R     & 972    & 825   & 337    & 1,566  & 957    & 996    & 51  & 5,704   \\
I     & 8,958  & 2,363 & 7,789  & 8,355  & 4,429  & 6,697  & 261 & 38,852  \\ \hline
Total & 25,075 & 9,324 & 23,598 & 27,741 & 13,609 & 18,886 & 713 & 118,946 \\ \hline
\end{tabular}%
}
\caption{Number of sentences by class and sector}
\label{datasets_statics}
\vskip -0.1in
\end{table*}
\begin{table*}[t]
\centering
\vskip 0.1in
\resizebox{\textwidth}{!}{%
\begin{tabular}{clccccccc}
\hline
Task & \multicolumn{1}{c}{Model}              & Automobile     & Aviation       & Electronics    & Finance        & Heavy industry & Steel industry & ESG-Kor        \\ \hline
\multirow{5}{*}{\begin{tabular}[c]{@{}c@{}}Four-class\\ classification\end{tabular}} & KoBERT & 84.16 & 91.62 & 82.68 & 87.03       & 90.60 & 85.84       & 84.70 \\
     & DistilKoBERT       & 83.53          & 90.75          & 81.98          & 86.33          & 90.64          & 85.33          & 84.19          \\
     & KB-ALBERT          & 83.88          & 91.15          & 82.28          & 86.57          & 91.17          & 84.69          & 84.59          \\
     & KLUE-RoBERTa-base  & {\ul 84.35}    & {\ul 92.16}    & {\ul 82.97}    & {\ul 87.83}    & {\ul 91.43}    & {\ul 86.14}    & {\ul 85.27}    \\
     & KLUE-RoBERTa-large & \textbf{84.59} & \textbf{92.64} & \textbf{83.82} & \textbf{88.28} & \textbf{91.56} & \textbf{87.30} & \textbf{85.87} \\ \hline
\multirow{5}{*}{\begin{tabular}[c]{@{}c@{}}Binary\\ classification\end{tabular}}     & KoBERT & 84.70 & 91.42 & 84.57 & {\ul 88.71} & 90.95 & {\ul 85.88} & 86.11 \\
     & DistilKoBERT       & 84.36          & 90.56          & 84.24          & 87.53          & 92.67          & 84.73          & 85.20          \\
     & KB-ALBERT          & 84.20          & 90.38          & 84.08          & 87.83          & 92.08          & 83.84          & 85.69          \\
     & KLUE-RoBERTa-base  & {\ul 85.29}    & {\ul 92.07}    & {\ul 85.48}    & 88.60          & {\ul 93.16}    & 85.23          & {\ul 86.30}    \\
     & KLUE-RoBERTa-large & \textbf{85.96} & \textbf{92.51} & \textbf{86.12} & \textbf{88.78} & \textbf{93.54} & \textbf{86.10} & \textbf{86.67} \\ \hline
\end{tabular}%
}
\caption{Classification accuracy (\%) by sector and model}
\label{Cls_results}
\vskip -0.1in
\end{table*}

\subsection{Exploratory Data Analysis}
\begin{figure}[h!]
\vskip 0.1in
\centering
\includegraphics[width=\linewidth]{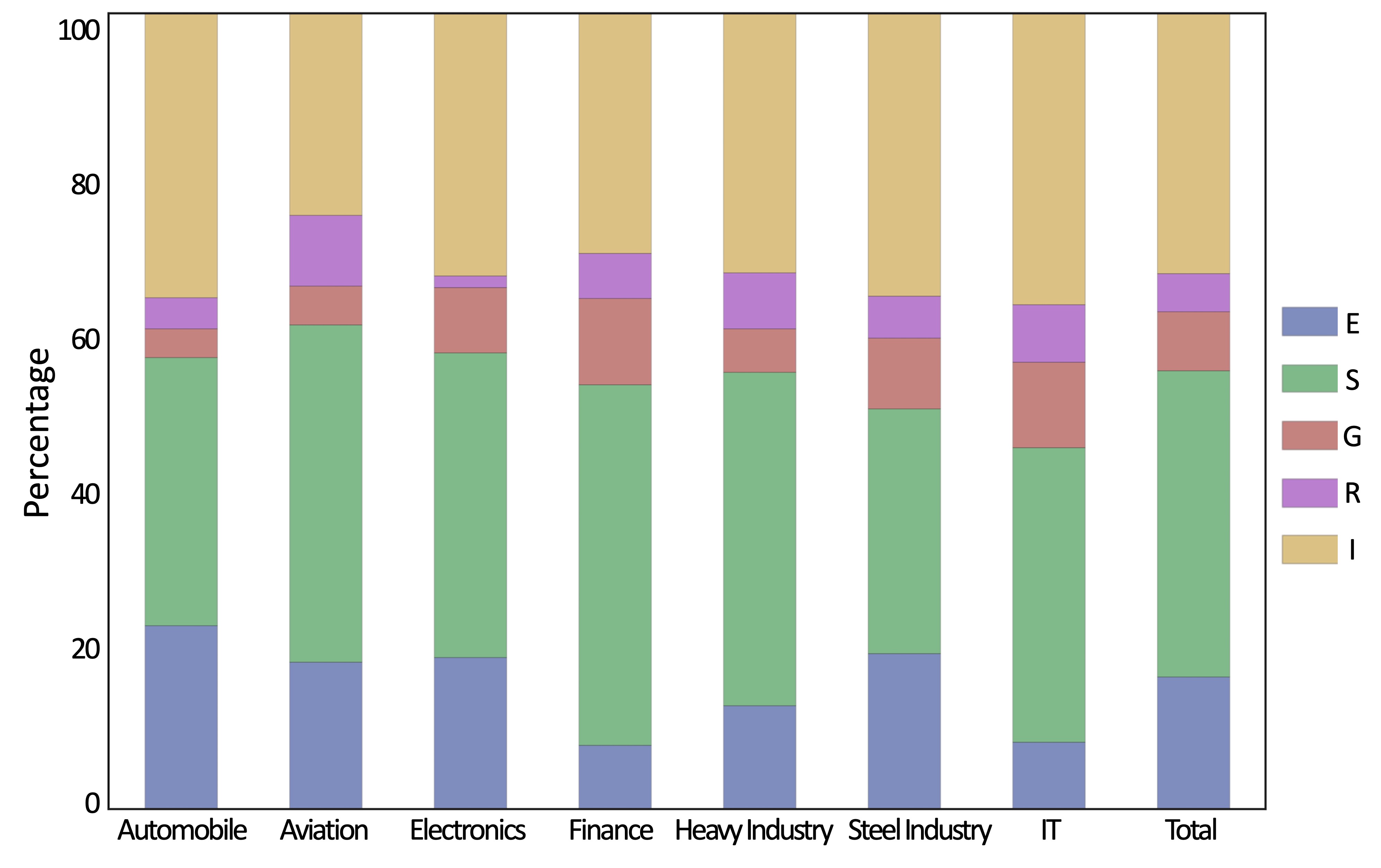}
\caption{Class distribution of classes by sector}
\label{dist_classes}
\vskip -0.1in
\end{figure}

After data preprocessing and labeling, the \texttt{ESG-Kor} dataset consists of a total of 118,946 sentences and classes, and it provides information about the year of issue and issuing company. Figure \ref{dist_classes} shows the distribution of labeled classes for each sector and the entire dataset.

As shown in Figure \ref{dist_classes}, most sectors show a similar distribution to the entire dataset. However, the finance sector has a significantly higher proportion of S class and fewer sentences labeled as E class than G class. The finance sector has less environmentally impactful activities and emphasizes social activities more than other industrial sectors. This confirms that the \texttt{ESG-Kor} dataset has been appropriately labeled according to real-world situations.

\begin{figure}[h]
\vskip 0.1in
\centering
\includegraphics[width=\linewidth]{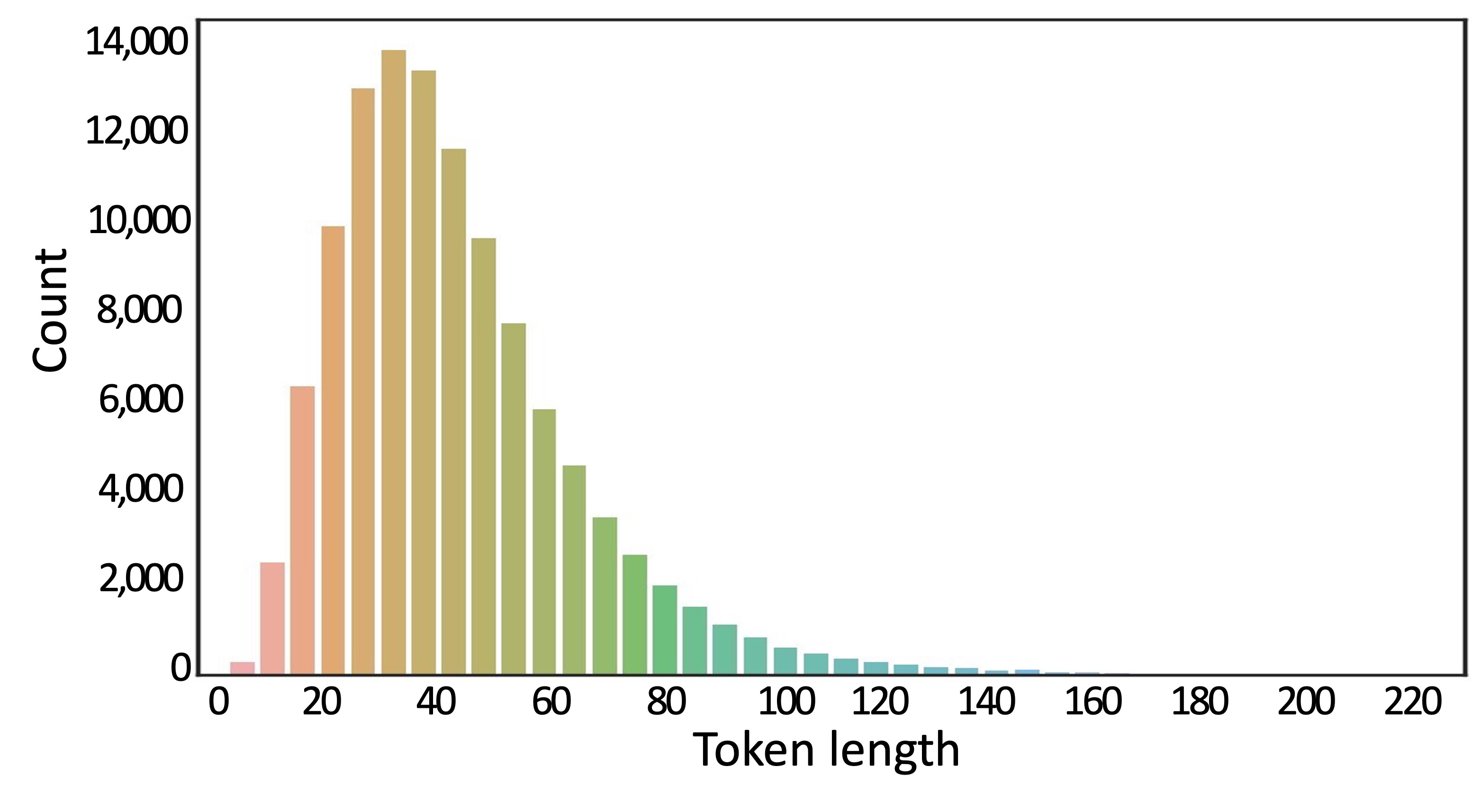}
\caption{Distribution of token length}
\label{dist_length}
\vskip -0.1in
\end{figure}

Since pre-trained language models are affected by the maximum token length of text data, the distribution of token lengths for each sentence in the \texttt{ESG-Kor} dataset is shown in Figure \ref{dist_length}.

The maximum token length of \texttt{ESG-Kor} is 233, the minimum token length is 3, the average length is 42, and the standard deviation is 20. 
The shortest sentences in the dataset consist of only three tokens and represent financial items such as \textit{current bad debt allowance}, \textit{cash and cash equivalents}, and \textit{current depreciation allowance}. These should be labeled as \texttt{Irrelevant} to clearly distinguish them from ESG information. After this sentence labeling, we manually reviewed and removed sentences that were too short or too long.
As most PLM models set the maximum token length to 512 or more, all sentences in \texttt{ESG-Kor} can be used as input without any tokens being cut off.

\section{Experiments}
\subsection{Experimental Settings}
To validate the effectiveness and applicability of \texttt{ESG-Kor}, we used models based on BERT \cite{devlin2019bert}. Since \texttt{ESG-Kor} is a Korean dataset, we utilized popular Korean pre-trained language models such as KoBERT \cite{SKTBrain2019korean}, DistilKoBERT \cite{park2019distilkobert}, KB-ALBERT \cite{kb2021kbalbert}, and KLUE-RoBERTa \cite{park2021klue}. We fine-tuned these pre-trained models using the \texttt{ESG-Kor} dataset.
The sentence $x$ of the \texttt{ESG-Kor} dataset is represented by the [CLS] token through the layers of the model, and the model calculates the probability of belonging to classes $c$ using [CLS] token $C$ and classification layer $W$, as shown in Equation \ref{eq1}. The model is trained to minimize the cross-entropy loss, as shown in Equation \ref{eq2}.
\begin{equation}
p(y=c) = \textit{softmax}(C W^{T})
\label{eq1}
\end{equation}
\begin{equation}
\mathcal{L} = -\sum_{c} y_{c} \log p(y=c)
\label{eq2}
\end{equation}
We set the batch size to \texttt{64}, maximum length to \texttt{256}, optimizer to \texttt{Adamw}, epsilon to \texttt{1e-08}, and the learning rate of KoBERT, DistilKoBERT, KB-ALBERT, and KLUE-RoBERTa to \texttt{1e-04}, \texttt{5e-04}, \texttt{5e-05}, and \texttt{5e-05}, respectively. The dataset was split into \texttt{70\%} training, \texttt{15\%} validation, and \texttt{15\%} test data. All experiments were conducted identically on \texttt{NVIDIA GeForce RTX 3090} 4ea.
The model parameters and the time taken for each experiment are provided in the Appendix \ref{app:models}.

\subsection{Classification Result}
The \texttt{ESG-Kor} dataset can be utilized for multiclass and binary classification tasks. The multiclass classification task classifies sentences into each E, S, and G class, and the binary classification task determines whether the sentence is related to ESG or not. The E, S, G, and R classes were consolidated into a \texttt{relevant} class for the binary classification dataset. At the same time, the \texttt{irrelevant} data remains the same as in the multiclass classification dataset. We didn't conduct sector-specific experiments for the IT sector due to the small number of collected data, but it was included in the dataset for the overall performance evaluation. The performance was evaluated based on classification accuracy, and the results for each sector and the overall experiment are shown in Table \ref{Cls_results}.

The \texttt{ESG-Kor} dataset comprises five classes, but we experimented with a four-class classification task by excluding the R class, which contains both ESG information simultaneously. In the experimental results, KLUE-RoBERTa-large showed the highest accuracy of 85.87\%, and even the DistilKoBERT, which recorded the lowest performance, confirmed meaningful performance with an accuracy of 84.19\%. In the binary classification experiment results, KLUE-RoBERTa-large showed the best performance with an accuracy of 86.67\%. 
In the experiments, models with more parameters performed better. However, with fewer parameters, KB-ALBERT performed as well as or better than KoBERT and DistilKoBERT. This can be interpreted as KB-ALBERT, a financial specialized model pre-trained on financial documents, understanding ESG documents better despite having fewer parameters. The number of parameters for the models is provided in the Appendix \ref{app:models}.
These experimental results confirmed that if fine-tuning is performed using the \texttt{ESG-Kor} dataset on Korean pre-trained language models, classification tasks can be appropriately performed regardless of the type of pre-trained language models. In other words, the experiment confirmed that when a dataset is specialized in a specific domain within a particular language, the applicability of pre-trained language models can be extended.

\section{Use Cases}
We propose two potential and practical cases to demonstrate additional applications of the \texttt{ESG-Kor} dataset and model. The first shows the scalability of ESG in the SME environment using a model trained on the \texttt{ESG-Kor} dataset. The second case shows a method that utilizes the trained model and additional methodologies for in-depth analysis of corporate ESG based on ESG key issues.

\subsection{SMEs ESG Information Classification}
\label{subsec:SMEs ESG info cls}
Korean SMEs are classified based on their revenue or asset size. SMEs are smaller companies in terms of income or assets. They account for approximately 99\% of all Korean enterprises. The importance of ESG management, previously dominated by large enterprises, is now increasing in SMEs due to reasons such as supplier selection and government procurement. However, most SMEs face difficulties with ESG management. We extend and validate ESG research into the SME environment through a model trained on the \texttt{ESG-Kor} dataset. Using ESG reports published by SMEs, we applied our methodology in an environment different from large enterprises. Since SMEs typically face challenges in publishing ESG reports, and their number is limited, we crawled the 2022 ESG reports of three SMEs through searches. We constructed an SME ESG data classification validation dataset using the same preprocessing and labeling process as the \texttt{ESG-Kor} dataset. The dataset consisted of 403 sentences and, considering the amount of data, was set up for binary classification related to ESG relevance rather than multiclass. The test results and class ratios using the model trained with the \texttt{ESG-Kor} dataset are as shown in Table \ref{SMEs_results}.

\begin{table}[h!]
\centering
\vskip 0.1in
\resizebox{\columnwidth}{!}{%
\begin{tabular}{lccc}
\hline
                            & \multicolumn{2}{c}{Number of data} & \multirow{2}{*}{Accuracy} \\ \cline{2-3}
                            & Relevant & Irrelevant &  \\ \hline
SME's binary classification & 254      & 149        & 86.72\%    \\ \hline
\end{tabular}%
}
\caption{Classification results for SME's}
\label{SMEs_results}
\vskip -0.1in
\end{table}
The experimental results showed that the model trained with the \texttt{ESG-Kor} dataset achieved an accuracy of 86.72\% in classifying ESG reports from SMEs. This confirms that a model trained on data from large corporations can be applicable in the SME environment, validating our methodology's ability to be extended to SME settings. Utilizing our methodology, SMEs can extract and classify ESG data, serving as an auxiliary tool for SMEs facing difficulties in ESG management, information disclosure, and report writing. Furthermore, it can be utilized across various ESG areas by all stakeholders related to SMEs.

\subsection{Extraction of ESG Key Issues}
\label{subsec:Extraction of ESG key}
MSCI is a global financial services company. In the ESG sector, MSCI evaluates and analyzes companies and investment products and is one of the most representative ESG rating agencies. MSCI, a leading ESG rating agency, proposes key issues for ESG evaluation. Within the environmental domain, there are four main issues and 13 detailed issues; in the social domain, there are four main issues and 14 detailed issues; and in the governance domain, there are two main issues and six detailed issues, making a total of 10 main categories and 33 detailed issues. More details are in Appendix \ref{App_Esg key issue}. MSCI utilizes these elements to evaluate and score companies' ESG performances. As companies generate vast amounts of information, directly scoring or extracting relevant information based on ESG key issues from this bulk is challenging and inefficient. Therefore, we demonstrate an efficient method to extract and analyze ESG information based on ESG key issues by combining a model trained with the \texttt{ESG-Kor} and NLP methodologies.

The process of extracting ESG information based on ESG key issues proceeds as follows. First, a model trained with the \texttt{ESG-Kor} dataset classifies ESG-related sentences from sustainability reports. Using a sentence embedding model, these sentences are mapped into the same vector space as ESG key issues. 
This step first extracts ESG-related data, saves resources and costs for embedding and retrieval, and filters out irrelevant noise information.
Next, semantic search based on vector cosine similarity is employed to extract sentences most similar to each ESG key issue. For experimentation, we used sustainability reports from Hyundai Construction, a company in the construction sector not included in the \texttt{ESG-Kor} dataset, spanning from 2010 to 2020. The data from Hyundai Construction's sustainability reports consisted of 6,403 sentences, from which 4,588 ESG-related sentences were extracted using the model trained on the \texttt{ESG-Kor} dataset. As the sentence embedding model, we utilized KoSimCSE with RoBERTa as its backbone, the Korean version of SimCSE \cite{gao2021simcse}, which effectively learns sentence representations through contrastive learning. Using KoSimCSE, we extracted the top 10 sentences with the highest similarity to each detailed ESG key issue from the extracted ESG-related sentences in Hyundai Construction's sustainability reports. Next, the extracted sentences were combined based on the main key issue and reconstructed into the top 10 sentences. We included at least one sentence with the highest cosine similarity for each detail issue. This was because sentences extracted related to a particular detail issue could show high cosine similarity, leading to the possibility of extracting sentences pertaining only to one detail issue. For example, one sentence with the highest similarity to Carbon Emissions, Product Carbon Footprint, Financing Environmental Impact, and Climate Change Vulnerability within the Climate Change issue was included in the results, and a total of 10 sentences were compiled based on similarity within the same main category issue. Examples of the extracted sentences are shown in Appendix \ref{App_Human_Eval}. To evaluate this approach, we conducted a human evaluation with 10 participants, and the evaluation scale and results are presented in Table \ref{Human_eval}.

\begin{table}[h]
\centering
\vskip 0.1in
\small
\resizebox{1.0\columnwidth}{!}{%
\begin{tabular}{cl|c}

\hline
\multicolumn{2}{c|}{Issues}                                               & \begin{tabular}[c]{@{}c@{}}Score\end{tabular} \\ \hline
\multicolumn{1}{c|}{} & Climate Change         & 4.9 \\ \cline{2-3} 
\multicolumn{1}{c|}{\multirow{2}{*}{Environmental}} & Natural Capital      & 4.6                                                        \\ \cline{2-3} 
\multicolumn{1}{c|}{} & Pollution \& Waste     & 4.7 \\ \cline{2-3} 
\multicolumn{1}{c|}{} & Env. opportunities     & 4.8 \\ \hline
\multicolumn{1}{c|}{} & Human Capital          & 4.8 \\ \cline{2-3} 
\multicolumn{1}{c|}{\multirow{2}{*}{Social}}        & Product Liability    & 4.5                                                        \\ \cline{2-3} 
\multicolumn{1}{c|}{} & Stakeholder Opposition & 4.6 \\ \cline{2-3} 
\multicolumn{1}{c|}{} & Social Opportunities   & 4.6 \\ \hline
\multicolumn{1}{c|}{\multirow{2}{*}{Governance}}    & Corporate Governance & 4.8                                                        \\ \cline{2-3} 
\multicolumn{1}{c|}{} & Corporate Behavior     & 4.4 \\ \hline
\end{tabular}%
}
\caption{The results of human evaluation. The evaluation scale is set as follows: 1 point = Not included, 2 points = Barely included, 3 points = Partially included, 4 points = Mostly included, and 5 points = Fully included.}
\label{Human_eval}
\vskip -0.1in
\end{table}

The human evaluation results indicated that the Climate Change issue within the Environmental category scored the highest, with a score of 4.9. This confirms that the extracted sentences appropriately included content related to Carbon Emissions, Product Carbon Footprint, Financing Environmental Impact, and Climate Change Vulnerability within Hyundai Construction's sustainability reports. The Governance Behavior issue scored the lowest, yet it still achieved a score of 4.4, indicating that the sentences extracted mostly contained content relevant to the issue. The average score across all evaluations was 4.67, validating our approach in appropriately extracting sentences related to all ESG key issues. 

Utilizing the presented method, it becomes feasible to easily access information related to required ESG issues amidst the vast and diverse information generated daily. It significantly facilitates extracting information related to ESG key issues from large text volumes, enabling more accurate and detailed analysis. Furthermore, as many ESG rating agencies base their ESG scores on key issues, extracting information relevant to these key issues can provide valuable insights for ESG scoring. In the future, the information extracted could contribute to the automation of ESG scoring for companies, enhancing the efficiency and accuracy of ESG evaluations.

\subsection{Discussion}
\texttt{ESG-Kor} can be utilized more diversely beyond the use cases presented in Section \ref{subsec:SMEs ESG info cls} and \ref{subsec:Extraction of ESG key}. Companies can use it to monitor their external ESG performance and compile sustainability reports by extracting or classifying ESG information from various sources. It can also be used internally to evaluate data for sustainability reports prepared for external disclosure. Individual investors or investment firms can extract and classify ESG information of specific companies for investment decision-making. Government and regulatory bodies can use the extracted and classified information to supervise companies and develop relevant policies. \texttt{ESG-Kor} can also be used for ESG scoring. Many agencies evaluate corporate ESG, but each has different criteria, leading to inconsistent ESG ratings for the same company. Additionally, these agencies do not disclose precise ESG evaluation criteria. Therefore, an automatic and objective ESG scoring system is needed, and \texttt{ESG-Kor} can be used to extract objective ESG information for scoring. Besides practical applications, \texttt{ESG-Kor} is also academically significant. Many researchers are currently conducting studies on ESG data analysis, ESG rating prediction, and the correlation between ESG and financial performance. However, available ESG-related datasets are often small or limited in application. Our dataset is expected to benefit ESG researchers in various fields.

\section{Conclusions}
We manually constructed the high-quality, large-scale ESG-related \texttt{ESG-Kor} dataset to effectively utilize pre-trained models in the ESG field to extract and classify ESG information. We collected and preprocessed sustainability reports containing ESG information to construct the dataset and annotated them manually according to various ESG criteria. The \texttt{ESG-Kor} dataset consists of 118,946 sentences, focusing on the automobile, aviation, electronics, finance, heavy industry, steel industry, and IT sectors. We conducted experiments using Korean pre-trained language models to verify the effectiveness and applicability of the \texttt{ESG-Kor} dataset. As a result of the experiments, the KLUE-RoBERTa-large model recorded accuracies of 85.87\% in multiclass classification and 86.67\% in binary classification, confirming that the model appropriately learned and performed the task. This suggests that applying the \texttt{ESG-Kor} dataset to existing pre-trained language models allows for proper extraction and classification of relevant information in the field of ESG of Korea, a specialized domain in a particular country. We also show use cases of how our methodology can be extended to the SME environment and the extraction of ESG key issues. By collecting SME ESG reports and labeling them, we evaluated the performance through a model trained with the \texttt{ESG-Kor} dataset, achieving an accuracy of 86.72\%. Furthermore, by combining our model with NLP techniques to extract ESG information based on ESG key issues, we validated its usefulness and potential in the ESG field through human evaluation.

In countries that use minority languages, research in specialized domains like ESG is limited due to the lack of datasets. Many datasets for minority languages are constructed by translating major languages, but translating specialized domains is challenging, and its performance is limited. Various countries are currently publishing documents such as sustainability reports that contain ESG information. Therefore, the entire construction process of the \texttt{ESG-Kor} dataset can serve as a benchmark for building ESG datasets in various countries. Furthermore, our approach can contribute to various ESG stakeholders, including researchers and practitioners, and can be utilized in multiple countries.

\section*{Limitation}
In preparation for future work, we analyzed the limitations of our research. First, our study focuses on classifying ESG information from text data. Table or graph data within sustainability reports also contain crucial information about each company's ESG. \texttt{ESG-Kor} was built using only the text within sustainability reports. We plan to expand the dataset to a broader scope by adding more diverse information, such as table and graph data. Second, our studies are based on BERT models. While it would have been possible to use large language models (LLMs) like Llama \cite{touvron2023llama}, which show excellent performance across various NLP tasks, we opted for models that were more accessible in terms of resources. Small language models can be more advantageous and easier to use in specific domain tasks or situations than large language models \cite{bosley2023we}. Additionally, our research primarily deals with corporate data, where LLMs may pose limitations regarding data privacy when used internally by corporations. Lastly, our work is currently limited to extracting and classifying ESG information. There might be a need for automatic scoring of ESG levels using the extracted ESG-related information. However, scoring requires the prior step of extracting and classifying ESG information from large texts, and the extracted ESG information can be directly utilized in various ways. The task of ESG information extraction is essential and meaningful, and we plan to extend it into a framework that can use the extracted ESG information to assign scores.

\section*{Ethics Statement}
The primary objective of this study is to enable ESG stakeholders to effectively extract and classify ESG-related information, thereby contributing positively to the ESG field and society. The dataset we have constructed does not contain any socially controversial or ethically inappropriate data, and there are no concerns related to the ethics statement in this study.

\section*{Acknowledgments}
This work was partly supported by Institute of Information \& communications Technology Planning \& Evaluation (IITP) grant funded by the Korea government (MSIT) (No.RS-2020-II201373, Artificial Intelligence Graduate School Program (Hanyang University)) and the research fund of Hanyang University (HY-202400000001018)

\bibliography{custom}

\newpage
\appendix
\setcounter{table}{0}
\renewcommand{\thetable}{\Alph{section}.\arabic{table}}

\section{Appendix}
\subsection{Model Details}
\label{app:models}
Table \ref{tab:model_params_time} provides the number of parameters for each model, along with the elapsed time for both the 4-class and 2-class classification tasks.
\begin{table}[h]
\centering
\resizebox{\columnwidth}{!}{%
\begin{tabular}{l|c|c|c}
\hline
Model & Number of & 4-class & 2-class \\
 & Parameters & Elapsed Time (s) & Elapsed Time (s) \\
\hline
KoBERT & 92M & 2812 & 2944 \\
DistilKoBERT & 28M & 809 & 839 \\
KB-ALBERT & 9M & 1409 & 1424 \\
KLUE-RoBERTa & 111M & 3133 & 3286 \\
KLUE-RoBERTa-large & 337M & 9492 & 9965 \\
\hline
\end{tabular}
}
\caption{Model parameters and elapsed time for experiments}
\label{tab:model_params_time}
\end{table}
\subsection{MSCI's ESG key issues}
\label{App_Esg key issue}
We utilized the ESG key issues from MSCI, a leading ESG rating agency, to automatically extract information related to them. MSCI presents key issues for evaluating each domain of ESG. MSCI's ESG key issues are first categorized into environmental, social, and governance domains and then further into related issues. These are subsequently broken down into detailed issues. The key issues for each ESG domain are shown in Table \ref{ESG_key}.

\begin{table}[h!]
\vskip 0.1in
\resizebox{\columnwidth}{!}{%
\begin{tabular}{cll}
\hline
\multicolumn{3}{c}{Issue} \\ \hline
\multicolumn{1}{c|}{} &
  \multicolumn{1}{l|}{Climate Change} &
  \begin{tabular}[c]{@{}l@{}}Carbon Emissions\\ Product Carbon Footprint\\ Financing Environmental Impact\\ Climate Change Vulnerability\end{tabular} \\ \cline{2-3} 
\multicolumn{1}{c|}{\multirow{2}{*}{Environmental}} &
  \multicolumn{1}{l|}{Natural Capital} &
  \begin{tabular}[c]{@{}l@{}}Water Stress\\ Biodiversity \& Land Use\\ Raw Material Sourcing\end{tabular} \\ \cline{2-3} 
\multicolumn{1}{c|}{} &
  \multicolumn{1}{l|}{Pollution \& Waste} &
  \begin{tabular}[c]{@{}l@{}}Toxic Emissions \& Waste\\ Packaging Material \& Waste\\ Electronic Waste\end{tabular} \\ \cline{2-3} 
\multicolumn{1}{c|}{} &
  \multicolumn{1}{l|}{Env. opportunities} &
  \begin{tabular}[c]{@{}l@{}}Clean Tech\\ Green Building\\ Renewable Energy\end{tabular} \\ \hline
\multicolumn{1}{c|}{} &
  \multicolumn{1}{l|}{Human Capital} &
  \begin{tabular}[c]{@{}l@{}}Labor Management\\ Health \& Safe\\ Human Capital Development\\ Supply Chain Labor Standards\end{tabular} \\ \cline{2-3} 
\multicolumn{1}{c|}{\multirow{2}{*}{Social}} &
  \multicolumn{1}{l|}{Product Liability} &
  \begin{tabular}[c]{@{}l@{}}Product safety \& Quality\\ Consumer Financial Protection\\ Privacy \& Data Security\\ Responsible Investment\\ Chemical safety\end{tabular} \\ \cline{2-3} 
\multicolumn{1}{c|}{} &
  \multicolumn{1}{l|}{Stakeholder Opposition} &
  \begin{tabular}[c]{@{}l@{}}Controversial Sourcing\\ Community Relations\end{tabular} \\ \cline{2-3} 
\multicolumn{1}{c|}{} &
  \multicolumn{1}{l|}{Social Opportunities} &
  \begin{tabular}[c]{@{}l@{}}Access to Finance\\ Access to Health Care\\ Opportunities in Nutrition \& Health\end{tabular} \\ \hline
\multicolumn{1}{c|}{\multirow{2}{*}{Governance}} &
  \multicolumn{1}{l|}{Corporate Governance} &
  \begin{tabular}[c]{@{}l@{}}Board\\ Pay\\ Ownership\\ Accounting\end{tabular} \\ \cline{2-3} 
\multicolumn{1}{c|}{} &
  \multicolumn{1}{l|}{Corporate Behavior} &
  \begin{tabular}[c]{@{}l@{}}Business Ethics\\ Tax Transparency\end{tabular} \\ \hline
\end{tabular}%
}
\caption{MSCI's ESG key issues}
\label{ESG_key}
\vskip -0.1in
\end{table}

\subsection{Human Evaluation}
\label{App_Human_Eval}
To evaluate each piece of information extracted based on ESG key issues, we conducted a human evaluation with 10 participants. To ensure accurate assessment, we scored the presence of detailed items under ESG key issues within a total of 10 pieces of extracted information in each domain on a scale of 1 (Not included) to 5 (Fully included). Examples of actual templates for each domain containing the ten pieces of extracted information are shown in Tables \ref{Humaneval_E}, \ref{Humaneval_S}, and \ref{Humaneval_G}. The extracted sentences in the examples were taken from the sustainability report of Hyundai Construction, a South Korean construction company, and translated from Korean to English.
다.

Table \ref{Humaneval_E} is the human evaluation template for Natural Capital in the environmental domain. Natural Capital consists of three detailed issues: water stress, biodiversity \& land use, and raw material sourcing. The extracted information includes content such as water usage at construction sites, rainwater utilization systems, analysis of the impact on the ecosystem around construction sites, and wastewater recycling, confirming the appropriate extraction of relevant information for Natural Capital. Table \ref{Humaneval_S} is the evaluation template for Human Capital in the social domain. Human Capital comprises four detailed issues: labor management, health \& safety, human capital development, and supply chain labor standards. This result includes content on building a talent development system, nurturing talents, institutional improvement through labor-management councils, improving the working environment, and reflecting special terms when selecting suppliers, confirming the appropriate extraction of Human Capital issues. Table \ref{Humaneval_G} provides an example of Corporate Behavior in the governance domain. Corporate Behavior includes two detailed issues: business ethics and tax transparency. The extracted information contains content on ethical management, compliance with fair trade, faithful tax payment, transparent disclosure of tax-related matters, and establishment of an ethical corporate culture. Corporate Behavior received the lowest human evaluation score of 4.4. This is likely because the content related to business ethics was relatively abstract, making it difficult for human evaluators to determine if it adequately addressed business ethics. However, it was confirmed that the extracted sentences included some content on ethical management. Thus, it can be interpreted that the current report does not deal with Business Ethics information in detail. This result can be extended for use in developing future ESG scoring systems.

\begin{table*}[b]
\begin{tabular}{|>{\centering\arraybackslash}m{3cm}|>{\centering\arraybackslash}m{12cm}|}
\hline
\texttt{Question} &
Do the following sentences include issues such as Natural Capital 

(Water Stress / Biodiversity \& Land Use / Raw Material Sourcing)? \\ \hline
\texttt{Extracted sentences} &
\begin{itemize}
    \item The necessary water is obtained on construction sites through surface water, groundwater, and tap water. 

    \item The impacts on the ecosystem surrounding the construction site, including biodiversity, are scientifically analyzed and incorporated into the biodiversity management strategy's planning and implementation process.

    \item Green purchasing involves buying products that contribute to resource savings and reduce environmental pollution compared to other products, services, or raw materials for the same use.

    \item Additionally, purchasing materials directly on-site reduces costs, enhances cost competitiveness, and increases the sales of local partners.

    \item A large amount of the water used at the site is for spraying to reduce fugitive dust, managing water usage and wastewater treatment.

    \item Environmentally friendly water use has become significant as the global water shortage intensifies, alongside energy, highlighting the reduction of water consumption in building operations.

    \item Rainwater utilization systems are applied to save water consumed during building operations by actively using rainwater for landscaping purposes.

    \item From the construction planning stage, the impacts on the ecosystem surrounding the construction site are scientifically analyzed to preserve biodiversity as much as possible and minimize changes in topography and vegetation damage by applying appropriate design and construction methods.

    \item On construction sites, water consumption is reduced by using rainwater storage facilities, groundwater dewatering systems, and sprinklers instead of water tank trucks for dust control.

    \item Additionally, the application of middle water recycling systems that recycle domestic sewage and rainwater utilization systems is being expanded on-site, and facilities that purify river water, seawater, and groundwater for environmentally friendly use are being established to optimize water resource usage.
\end{itemize}
\\ \hline
\texttt{Inclusion level} &
\begin{tabular}{@{}l@{}}
$\bigcirc$ Not included \\[\baselineskip]
$\bigcirc$ Barely included \\[\baselineskip]
$\bigcirc$ Partially included \\[\baselineskip]
$\bigcirc$ Mostly included \\[\baselineskip]
$\bigcirc$ Fully included
\end{tabular}
 \\ \hline
\end{tabular}
\caption{Example of an evaluation template for the environmental domain for human evaluation}
\label{Humaneval_E}
\end{table*}

\begin{table*}[b]
\begin{tabular}{|>{\centering\arraybackslash}m{3cm}|>{\centering\arraybackslash}m{12cm}|}
\hline
\texttt{Question} &
Do the following sentences include issues such as Human Capital 

(Labor Management / Health \& Safe / Human Capital Development / Supply Chain Labor Standard)? \\ \hline
\texttt{Extracted sentences} &
\begin{itemize}
    \item Additionally, we are developing experts by establishing an advanced talent development system through the eHRD (Human Resource Development) system.

    \item Continuous talent cultivation is necessary for sustained growth, requiring the nurturing of excellent personnel.

    \item The labor union discusses system improvements, productivity enhancement, welfare, and working environment improvements through the labor-management council.

    \item By establishing safety and health policies aimed at instilling a safety culture through education and improving worker welfare, we regularly track and manage risks related to occupational health and safety, providing employees with a safe working environment.

    \item Based on this, we include environmental, labor, and health and safety standards as special conditions in supplier selection contracts, making them mandatory for all contracts.

    \item To this end, we are implementing talent development strategies, presenting a vision for personal growth, and supporting the development of global competencies.

    \item Additionally, we conduct leadership training and operate global advanced study programs to develop our employees into global leaders.

    \item We are expanding investments in employee competency development, striving to cultivate multifaceted talents with rich creativity and imagination, beyond simple problem-solving abilities.

    \item Hyundai Engineering \& Construction is making continuous efforts for talent development through domestic and international training programs and in-house online education.

    \item We plan to strengthen immediately applicable training programs, such as CPD (Continuing Professional Development) for engineers, competency enhancement courses for site managers, and leadership training for those promoted, while also expanding education for employees' spouses and children.
\end{itemize}
\\ \hline
\texttt{Inclusion level} &
\begin{tabular}{@{}l@{}}
$\bigcirc$ Not included \\[\baselineskip]
$\bigcirc$ Barely included \\[\baselineskip]
$\bigcirc$ Partially included \\[\baselineskip]
$\bigcirc$ Mostly included \\[\baselineskip]
$\bigcirc$ Fully included
\end{tabular}
 \\ \hline
\end{tabular}
\caption{Example of an evaluation template for the social domain for human evaluation}
\label{Humaneval_S}
\end{table*}

\begin{table*}[b]
\begin{tabular}{|>{\centering\arraybackslash}m{3cm}|>{\centering\arraybackslash}m{12cm}|}
\hline
\texttt{Question} &
Do the following sentences include issues such as Corporate Behavior 

(Business Ethics / Tax Transparency)? \\ \hline
\texttt{Extracted sentences} &
\begin{itemize}
    \item Practicing ethical management is not a choice but a necessity for a company's sustainable growth.

    \item Hyundai Engineering \& Construction maintains an open, honest, and positive relationship with tax authorities, faithfully pays taxes annually according to each country's policies, and is committed to transparently disclosing all tax-related matters.

    \item We are devoting our entire capabilities to realizing the values of sustainable management, such as transparent ethical management, environmentally friendly business operations, win-win development with partners, and fulfilling social responsibilities.

    \item Expanding the scope of application, Hyundai Engineering \& Construction practices ethical management together with stakeholders related to its business.

    \item Hyundai Engineering \& Construction maintains an open, honest, and positive relationship with tax authorities, including the National Tax Service, and faithfully pays taxes annually according to each country's tax policies.

    \item Hyundai Engineering \& Construction is leading the establishment of an ethical corporate culture not only among its employees but also among its affiliates, partners, and business associates.

    \item All employees and partners of Hyundai Engineering \& Construction share and internalize the company's comprehensive ethical management system, practicing the code of ethics that reflects long-term corporate value.

    \item As a responsible corporate citizen, Hyundai Engineering \& Construction strives to become a respected company among stakeholders through ethical business activities and fair trade compliance.

    \item Through this, the company has demonstrated its strong commitment to not only ethical management but also fulfilling its social responsibilities.

    \item The global market, where top-tier companies compete, is our stage of activity.
\end{itemize}
\\ \hline
\texttt{Inclusion level} &
\begin{tabular}{@{}l@{}}
$\bigcirc$ Not included \\[\baselineskip]
$\bigcirc$ Barely included \\[\baselineskip]
$\bigcirc$ Partially included \\[\baselineskip]
$\bigcirc$ Mostly included \\[\baselineskip]
$\bigcirc$ Fully included
\end{tabular}
 \\ \hline
\end{tabular}
\caption{Example of an evaluation template for the governance domain for human evaluation}
\label{Humaneval_G}
\end{table*}

\end{document}